\relax
\documentclass[letterpaper]{article} 
\usepackage{aaai22}  
\usepackage{times}  
\usepackage{helvet}  
\usepackage{courier}  
\usepackage[hyphens]{url}  
\usepackage{graphicx} 
\usepackage{natbib}  
\usepackage{caption} 
\DeclareCaptionStyle{ruled}{labelfont=normalfont,labelsep=colon,strut=off} 
\usepackage{subfig}
\usepackage{float}
\usepackage{graphicx}

\usepackage{algorithm}
\usepackage{algorithmic}
\usepackage{bibentry}
\usepackage{multirow}
\usepackage{diagbox}
\usepackage{amssymb}
\usepackage{amsmath}
\usepackage{booktabs}
\usepackage{diagbox}
\usepackage[switch]{lineno}
\usepackage{newfloat}
\usepackage{listings}
\floatstyle{ruled}
\newfloat{listing}{tb}{lst}{}
\floatname{listing}{Listing}

\begin{document}

\title{Collaboration of Pre-trained Models Makes Better Few-shot Learner}

\author{
    \vspace{1pt}\\
    Renrui Zhang\textsuperscript{\rm 1,2}\equalcontrib, 
    Bohao Li\textsuperscript{\rm 3}\equalcontrib,
    Wei Zhang\textsuperscript{\rm 1}, 
    Hao Dong\textsuperscript{\rm 4}, 
    \\Hongsheng Li\textsuperscript{\rm 2}, 
    Peng Gao\textsuperscript{\rm 1}, 
    Yu Qiao\textsuperscript{\rm 1}\\
    \vspace{4pt}
}
\affiliations{
    
    \textsuperscript{\rm 1} Shanghai AI Laboratory, \textsuperscript{\rm 2} The Chinese University of Hong Kong,\\
    \textsuperscript{\rm 3} University of Chinese Academy of Sciences,
    \textsuperscript{\rm 4} Peking University\\\vspace{4pt}
    \{zhangrenrui, denghanqiu, gaopeng, qiaoyu\}@pjlab.org.cn\\


%
}

\maketitle

\begin{abstract}
Few-shot classification requires deep neural networks to learn generalized representations only from limited training images, which is challenging but significant in low-data regimes. Recently, CLIP-based methods have shown promising few-shot performance benefited from the contrastive language-image pre-training. Based on this point, we question if the large-scale pre-training can alleviate the few-shot data deficiency and also assist the representation learning by the pre-learned knowledge. In this paper, we propose \textbf{CoMo}, a \textbf{Co}llaboration of pre-trained \textbf{Mo}dels that incorporates diverse prior knowledge from various pre-training paradigms for better few-shot learning. Our CoMo includes: CLIP's language-contrastive knowledge, DINO's vision-contrastive knowledge, and DALL-E's language-generative knowledge. Specifically, CoMo works in two aspects: few-shot data expansion and diverse knowledge ensemble. For one, we generate synthetic images via zero-shot DALL-E to enrich the few-shot training data without any manpower. For the other, we introduce a learnable Multi-Knowledge Adapter (MK-Adapter) to adaptively blend the predictions from CLIP and DINO. By such collaboration, CoMo can fully unleash the potential of different pre-training methods and unify them to perform \textit{state-of-the-art} for few-shot classification. We conduct extensive experiments on 11 datasets to demonstrate the superiority and generalization ability of our approach.

\end{abstract}

\section{Introduction}
Convolutional neural networks~\cite{NIPS2012_c399862d} and transformers~\cite{NIPS2017_3f5ee243} have attained great success on a wide range of vision tasks, which is based on thorough training with abundant datasets~\cite{deng2009imagenet}. Instead, for some data-deficient and resource-finite scenarios, few-shot learning~\cite{NIPS2016_90e13578, snell2017prototypical} also becomes a research hotspot that learns a well-performed network under limited images with annotations. Many previous works have been proposed in this field to enhance model's generalization capability by meta learning~\cite{finn2017model, LearningToLearn16}, metric learning~\cite{zhang2020deepemd}, and data augmentation~\cite{Hallucinating17, yang2021free}. Recently, CLIP~\cite{radford2021learning} pre-trained by large-scale language-image pairs shows favorable zero-shot transfer ability for open-vocabulary visual recognition. The follow-up CoOp~\cite{coop}, CLIP-Adapter~\cite{2110.04544} and Tip-Adapter~\cite{zhang2021tip} further extend it for few-shot classification and achieve superior performance on various downstream datasets. This indicates that, even if the few-shot training data is insufficient, the large-scale pre-training has endowed the network with strong representation ability, which highly benefits the few-shot learning on downstream domains. Now that there exist various self-supervisory paradigms, could we adaptively integrate their pre-learned knowledge and collaborate them to be a better few-shot learner?

\begin{figure}[t]
\centering
\includegraphics[width=1\linewidth]{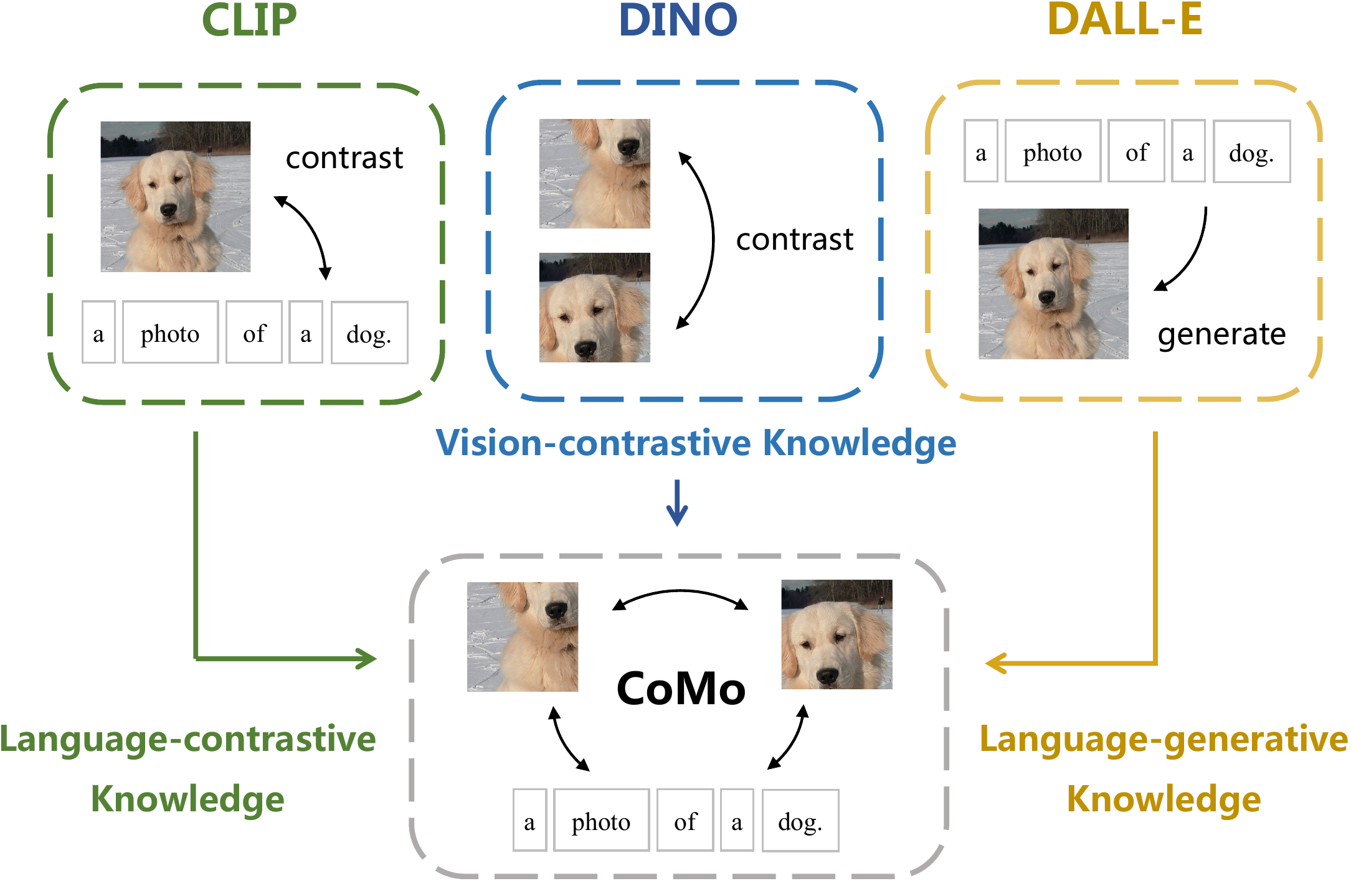}
\caption{\textbf{The Collaboration Paradigm of CoMo.} We adaptively incorporate the knowledge from three types of pre-training methods and achieve a strong few-shot learner.}
\label{fig1}
\end{figure}

To tackle this issue, we propose \textbf{CoMo}, a \textbf{Co}llaboration of \textbf{Mo}dels blending the knowledge from multiple pre-training paradigms for few-shot image classification. As shown in Figure~\ref{fig1}, we integrate CLIP~\cite{radford2021learning}, DINO~\cite{Caron_2021_ICCV}, and DALL-E~\cite{pmlr-v139-ramesh21a} to provide three types of prior knowledge for CoMo. Therein, CLIP~\cite{radford2021learning} is pre-trained to produce paired features in the embedding space for every image and its descriptive text. Guided by texts with different categorical semantics, CLIP~\cite{radford2021learning} can clearly classify the images and contain the \textbf{language-contrastive knowledge}. DINO follows contrastive self-supervised learning~\cite{Caron_2021_ICCV} to match the 
representations between two transformations of one same image, which is expert at distinguishing different images with \textbf{vision-contrastive knowledge}. Similar to CLIP~\cite{radford2021learning}, DALL-E~\cite{pmlr-v139-ramesh21a} is also pre-trained by image-text pairs but learns to predict the encoded image tokens based on the text tokens. Conditioned on the input text, DALL-E~\cite{pmlr-v139-ramesh21a} could leverage the \textbf{language-generative knowledge} to create high-quality synthetic images in a zero-shot manner. Therefore, the three models have distinctive pre-training goals and can provide complementary knowledge to assist the few-shot visual recognition.

In detail, we make them collaborate from two perspectives: few-shot data expansion and diverse knowledge ensemble. Firstly, we adopt DALL-E~\cite{pmlr-v139-ramesh21a} to generate additional training images for different categories based on the domain-specific texts, which enlarges the few-shot training data but costs no extra manpower for collection and annotation. Then, we propose a Multi-Knowledge Adapter (MK-Adapter) that adaptively incorporates the predictions from both CLIP~\cite{radford2021learning} and DINO~\cite{Caron_2021_ICCV}. Referring to Tip-Adapter~\cite{zhang2021tip}, we build a cache model but with two kinds of keys respectively for the two pre-trained models, which can produce two classification logits by vision-contrast for the input image. Regarding CLIP's zero-shot logits of language-contrast as the baseline, we reweight the two adapters' logits via distribution similarity and ensemble all the three logits as the final output. By fine-tuning only the lightweight MK-Adapter via expanded training data, CoMo can learn to effectively fuse the prior knowledge of the three types of pre-training and leverage their complementary characteristics for better few-shot visual recognition.

Our main contributions are summarized as follows:
\begin{enumerate}
    \item We propose CoMo to incorporate the prior knowledge learned from various pre-training paradigms for better few-shot learning. 
    \item By collaborating the pre-trained CLIP, DINO, and DALL-E, CoMo enriches the limited few-shot training data and adaptively ensembles diverse predictions via the MK-Adapter.
    \item We conduct thorough experiments on 11 datasets for few-shot classification, where our CoMo achieves \textit{state-of-the-art} performance without using extra annotated data.
\end{enumerate}

\section{Related Work}
\subsection{Pre-training of Visual Models}
With the breakthroughs in deep learning models~\cite{He_2016_CVPR,2010.11929,Liu_2021_ICCV}, most modern vision models are based on the paradigm of pre-training on ImageNet and fine-tuning on downstream tasks~\cite{He_2019_ICCV}. Pre-trained models have shown promising adaptability for various downstream tasks, such as fine-tuning classification~\cite{Kornblith_2019_CVPR}, object detection~\cite{Lin_2017_ICCV} and semantic segmentation~\cite{7913730}, etc. 

\begin{figure*}[t]
\centering
\includegraphics[width=1.0\textwidth]{ 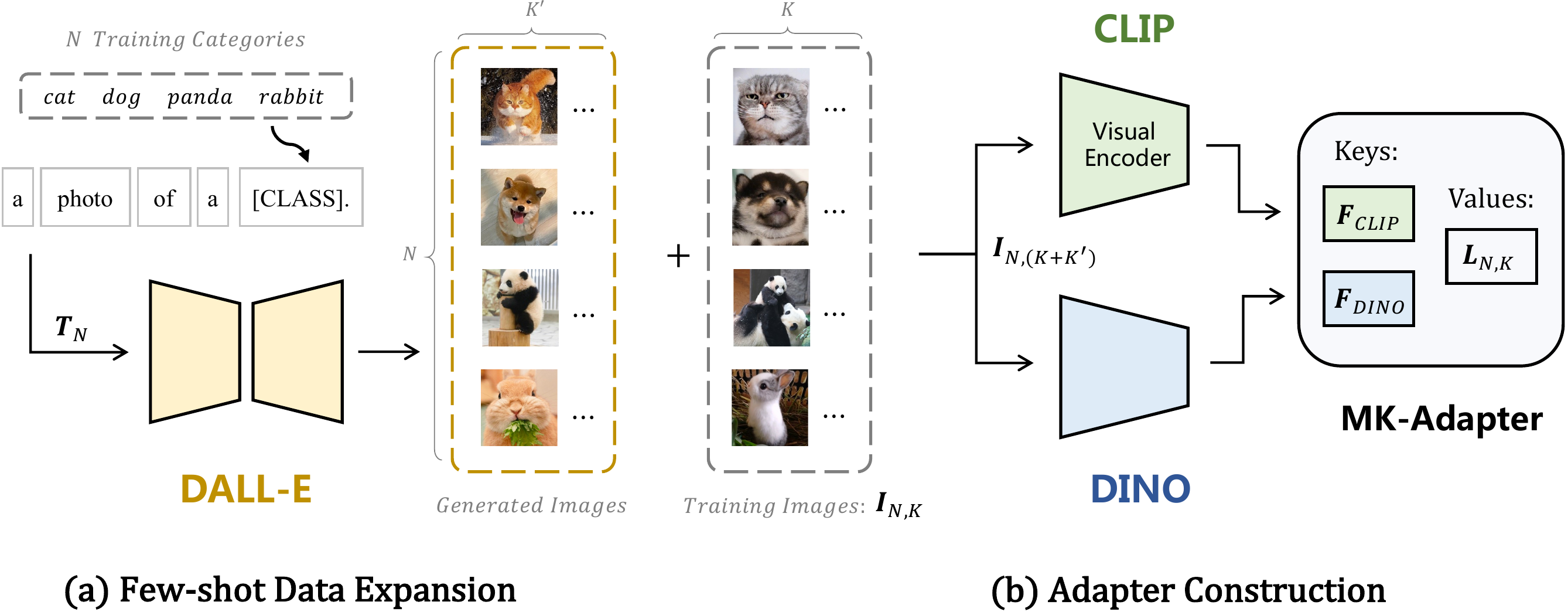} 
\caption{(a) We adopt DALL-E~\cite{pmlr-v139-ramesh21a} to generate synthetic images to expand the limited few-shot training samples. (b) We construct the Multi-Knowledge Adapter with two kinds of keys to adaptively fuse the knowledge from CLIP~\cite{radford2021learning} and DINO~\cite{Caron_2021_ICCV}.}
\label{fig:Framework1}
\end{figure*}

To improve the representation capability by overcoming the constraints of annotation, unsupervised/self-supervised pre-training has raised wide concerns by propagating from small annotated samples to large scale unlabeled datasets ~\cite{9086055}. Self-supervised learning is initialized as pretext tasks, such as image restoration from corruption~\cite{10.1145/1390156.1390294,Pathak_2016_CVPR,He_2022_CVPR}, pseudo labels~\cite{Doersch_2015_ICCV,10.1007/978-3-319-46466-4_5} and clustering~\cite{Caron_2018_ECCV}. Recently, contrast learning, which learns representations by contrasting positive pairs against negative pairs, has gotten well studied for diverse self-supervised visual representation learning~\cite{He_2020_CVPR,1807.03748,2003.04297,pmlr-v119-chen20j,NEURIPS2020_f3ada80d,Caron_2021_ICCV}. Besides, language-supervised visual pre-training emerges as a novel paradigm closer to natural visual understanding~\cite{Miech_2019_ICCV,1504.00032,sharma-etal-2018-conceptual,Antol_2015_ICCV}, among which CLIP~\cite{radford2021learning} obtains powerful zero-shot transferability by contrastive pre-training on image-text pairs collected from the Internet. In addition, vision-language pre-training can also promote the zero-shot image generation from text. Open generative models, such as DALL-E~\cite{pmlr-v139-ramesh21a} and CogView~\cite{NEURIPS2021_a4d92e2c} pre-trained on large scale image-text pairs are able to generate images with complex structures and diverse contents conditioned on a given text.\\

\subsection{Few-shot Adaptation}
Few-shot learning highly relies on the transferability of the trained neural networks. 
From the perspective of distance measurement, some metric learning methods learn a metric space by computing the distances from the instances to novel categories~\cite{snell2017prototypical,Sung_2018_CVPR,NIPS2016_90e13578}. Also, meta-learning is proposed to improve the few-shot adaptation ability of the models by finding a set of initialized parameters that can rapidly adapt to novel domains~\cite{pmlr-v70-finn17a,Jamal_2019_CVPR,Chen_2021_ICCV, li2021beyond}. More recently, with the vision-language pre-training model CLIP~\cite{radford2021learning} exhibiting strong zero-shot adaptation performance, several efforts have started to find efficient strategies to adapt it to downstream few-shot datasets. Following the recent research trend of NLP in prompt learning, context optimization (CoOp)~\cite{coop} is proposed as a prompt tuning adaptation method by optimizing a set of learnable prompt tokens. Subsequently, to address the harm of generalization ability brought by CoOp~\cite{coop}, that is, the recognition of unknown categories becomes poor, CoCoOp~\cite{Zhou_2022_CVPR} proposes to train a meta-network to generate image tokens as conditional inputs for the textual vectors. Further, referring to adapters~\cite{houlsby2019parameter} in NLP, CLIP-Adapter~\cite{2110.04544} is introduced to fine-tune CLIP by applying lightweight residual-style bottleneck layers as the adapter. Tip-Adapter~\cite{zhang2021tip} is then proposed as a training-free adaption method with a constructed key-value cache model. It can also be regarded as a better initialization of CLIP-Adapter with much faster convergence when fine-tuning. 

Our work proposes the collaboration of multiple pre-trained visual models as a unified framework and well leverage their strength to adapt CLIP for few-shot learning: the vision-language model CLIP~\cite{radford2021learning} for zero-shot image classification, the self-supervised visual transformer DINO~\cite{Caron_2021_ICCV} to provide distinctive visual representations, and the text-to-image generative model DALL-E~\cite{pmlr-v139-ramesh21a} for zero-shot data generation.

\section{Collaboration of Pre-trained Models}

In this section, we first briefly revisit three types of pre-training paradigms for CoMo. Then, we specifically introduce how our CoMo collaborates them for few-shot classification.

\subsection{Different Pre-training Paradigms}
\label{s3-1}

\paragraph{Contrastive Vision-Language Pre-training.}
The series~\cite{pmlr-v119-chen20j} of contrastive learning between vision and language learn to map the two modalities into the same embedding space via a contrastive loss. Driven by web-scale datasets, e.g., 400 million for CLIP~\cite{radford2021learning} and 1.8 billion for ALIGN~\cite{ALIGN}, the basic pre-training target is to minimize the embedding distances of images and their textual descriptions, while maximize those unpaired ones. By the cross-modal alignment, we can discriminate images of different categorizes by the texts with different semantics. We denote such learned prior as language-contrastive knowledge and adopt CLIP as the representative model for such pre-training method.

\paragraph{Contrastive Vision Pre-training.}
As the traditional self-supervised learning methods, vision-contrastive models~\cite{pmlr-v119-chen20j} focus on the discrimination between different images. Normally, the positive pairs to be drawn close are two transformations of the same image, while the optimization of negative pairs~\cite{grill2020bootstrap} is optional, which can be replaced by a momentum encoder~\cite{He_2020_CVPR} or cluster assignments~\cite{caron2020unsupervised}. Recent works reveal that we can learn self-supervised features without negative pairs between images~\cite{NEURIPS2020_f3ada80d,Caron_2021_ICCV}.
Given the strong linear classification capacity, the pre-trained DINO~\cite{Caron_2021_ICCV} is adopted here to provide vision-contrastive knowledge for collaboration.

\paragraph{Generative Vision-Language Pre-training.}
Learned from millions of image-caption pairs, the DALL-E series can generate language-conditioned images in a zero-shot manner. They are pre-trained to autoregressively predict the encoded image tokens from the textual tokens of the captions. With such language-generative knowledge, the pre-trained DALL-E can be viewed as a free lunch to enlarge the training data without any manpower. Considering publicity, we select DALL-E-mini~\cite{Mini_Dalle} as the representative among DALL-E models.


\begin{figure*}[t]
\centering
\includegraphics[width=1.0\textwidth]{ 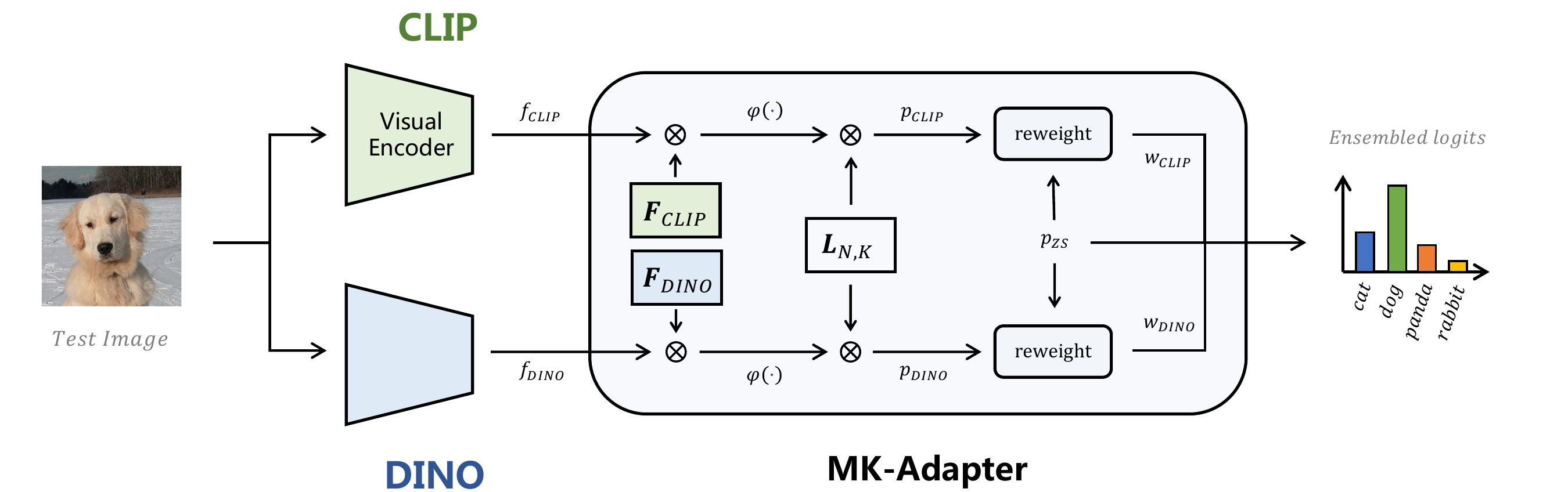} 
\caption{\textbf{Adaptive Ensemble of DINO and CLIP via Multi-knowledge Adapter.} We first regard the test image as the query and retrieves CLIP and DINO's knowledge from the corresponding two keys. Then, we calculate the distribution similarities between different classification logits for adaptive ensemble.}
\label{fig:Framework2}
\end{figure*}

\subsection{Few-shot Data Expansion}
\label{s3-2}

Under the $N$-way $K$-shot settings, we have the few-shot training images $I_{N,K}$ with labels $L_{N,K}$ that contain $K$ samples for each $N$ categories. Via the zero-shot DALL-E~\cite{Mini_Dalle}, we generate synthesis images to enrich our limited training data, as shown in Figure~\ref{fig:Framework1} (a). For different categories, we adopt a unified simple template, e.g., ``a photo of a [CLASS].'', and put the category name into [CLASS] as the input for DALL-E. After the generation, we utilize CLIP to filter the top-$K'$ best-quality images as the newly-expanded training samples for each category. Then, we obtain the $N$-category ($K+K'$)-sample training images, formulated as
\begin{align}
    I_{N,(K+K')}= \{\text{DALL-E}(T_{N}),\ \ I_{N,K}\},
   \label{eq:dalle},
\end{align}
where $T_N$ denotes the $N$-category textual inputs.
We keep $K'$ comparable with $K$ to ensure the synthesis quality and also preserve the low-data regimes. By the pre-trained language-generative knowledge, the data expansion is totally zero-shot, which not only requires no manpower to collect or annotate the data, but also alleviates the data deficiency issue inherently for few-shot learning nearly at no cost.

\subsection{Diverse Knowledge Ensemble.}
\label{s3-3}

Besides DALL-E for expanding few-shot data, we introduce a Multi-Knowledge Adapter, termed as MK-Adapter, which constructs a cache model to adaptively collaborate the pre-trained CLIP~\cite{radford2021learning} and DINO~\cite{Caron_2021_ICCV}. 

\paragraph{Adapter Construction.}
MK-Adapter adopts the cache-based adapter that retrieves knowledge from constructed key-value cache model. Different from Tip-Adapter~\cite{zhang2021tip} only adapting CLIP, our cache model contains the pre-learned knowledge from both CLIP and DINO by caching two kinds of keys. Specifically in Figure~\ref{fig:Framework2} (b), we first utilize CLIP and DINO to independently extract visual features of the few-shot training images, formulated as
\begin{align}
    F_{\text{CLIP}} &= \text{CLIP$_{vis}$}(I_{N,(K+K')});\\
    F_{\text{DINO}} &= \text{DINO}(I_{N,(K+K')}),
   \label{eq:extract}
\end{align}
where CLIP$_{vis}$ denotes the CLIP's visual encoder and $F_{\text{CLIP}}, F_{\text{DINO}}\in \mathbb{R}^{N(K+K')\times C}$. Besides the two keys, we convert the few-shot training labels into one-hot encodings $L_{N,(K+K')}\in \mathbb{R}^{N(K+K')\times N}$, and regard them as the only values for both keys. During training, we follow Tip-Adapter that only enables the cached keys in the adapter to be learnable and keeps the pre-trained models frozen.

\paragraph{Adaptive Ensemble.}
For a test image in Figure~\ref{fig:Framework2}, we first extract its two visual features $f_{\text{CLIP}},f_{\text{DINO}}\in \mathbb{R}^{1\times C}$ and regard them as queries to retrieve diverse knowledge from the MK-Adapter. Then, we could acquire three predicted classification logits $p_{\text{ZS}},p_{\text{CLIP}},p_{\text{DINO}}\in \mathbb{R}^{1\times N}$, which are respectively from CLIP's zero-shot textual alignment and the two keys of MK-Adapter. We formulate them as
\begin{align}
    p_{\text{ZS}} &= f_{\text{CLIP}}\ \text{CLIP$_{tex}$}(T_N)^T;\\
    p_{\text{CLIP}} &= \varphi(f_{\text{CLIP}}F_{\text{CLIP}}^T)\ L_{N,(K+K')};\\
    p_{\text{DINO}} &= \varphi(f_{\text{DINO}}F_{\text{DINO}}^T)\ L_{N,(K+K')},
   \label{eq:logits}
\end{align}
where CLIP$_{tex}$ represents CLIP's textual encoder, and $f_{\text{CLIP}}\&F_{\text{CLIP}}^T$ denotes the query-key affinity matrix of CLIP's adapter, analogous to DINO's. $\varphi(x) = \exp(-\beta\cdot(1-x))$ serves as a non-linear modulator to control the sharpness of affinity matrix.
As the language-contrastive $p_{\text{ZS}}$ is pre-trained by 400 million data and can perform strong zero-shot transfer ability, we regard $p_{\text{ZS}}$ as the prediction baseline and calculate the weights of $p_{\text{CLIP}},p_{\text{DINO}}$ for ensemble based on their distribution similarity with $p_{\text{ZS}}$. By this, we can suppress some obviously false category possibilities in $p_\text{CLIP},p_\text{DINO}$ and also amplify the moderately correct ones during ensemble. Firstly, we respectively normalize the scales of three classification logits into -1$\sim$1 by their each mean and standard deviation. We then calculate the distribution similarities as the ensemble weights for the two logits of the adapter as
\begin{align}
    w_{\text{CLIP}} = p_{\text{CLIP}}\ p_{\text{ZS}}^T;\ \ w_{\text{DINO}} = p_{\text{DINO}}\ p_{\text{ZS}}^T.
   \label{eq:sim}
\end{align}
Finally, we adopt the softmax function to normalize the weights and obtain the final ensemble logits as
\begin{align}
    p_{en} = p_{\text{ZS}} + \sum_{i}p_i\cdot\text{softmax($w_i$)},
   \label{eq:sim}
\end{align}
where $i\in \{\text{CLIP}, \text{DINO}\}$. By such similarity-based ensemble, $p_{en}$ can adaptively fuse the prior knowledge learned by CLIP and DINO's pre-training and achieve stronger few-shot image classification.

\begin{figure}[htb]
\centering
\includegraphics[width=0.9\linewidth]{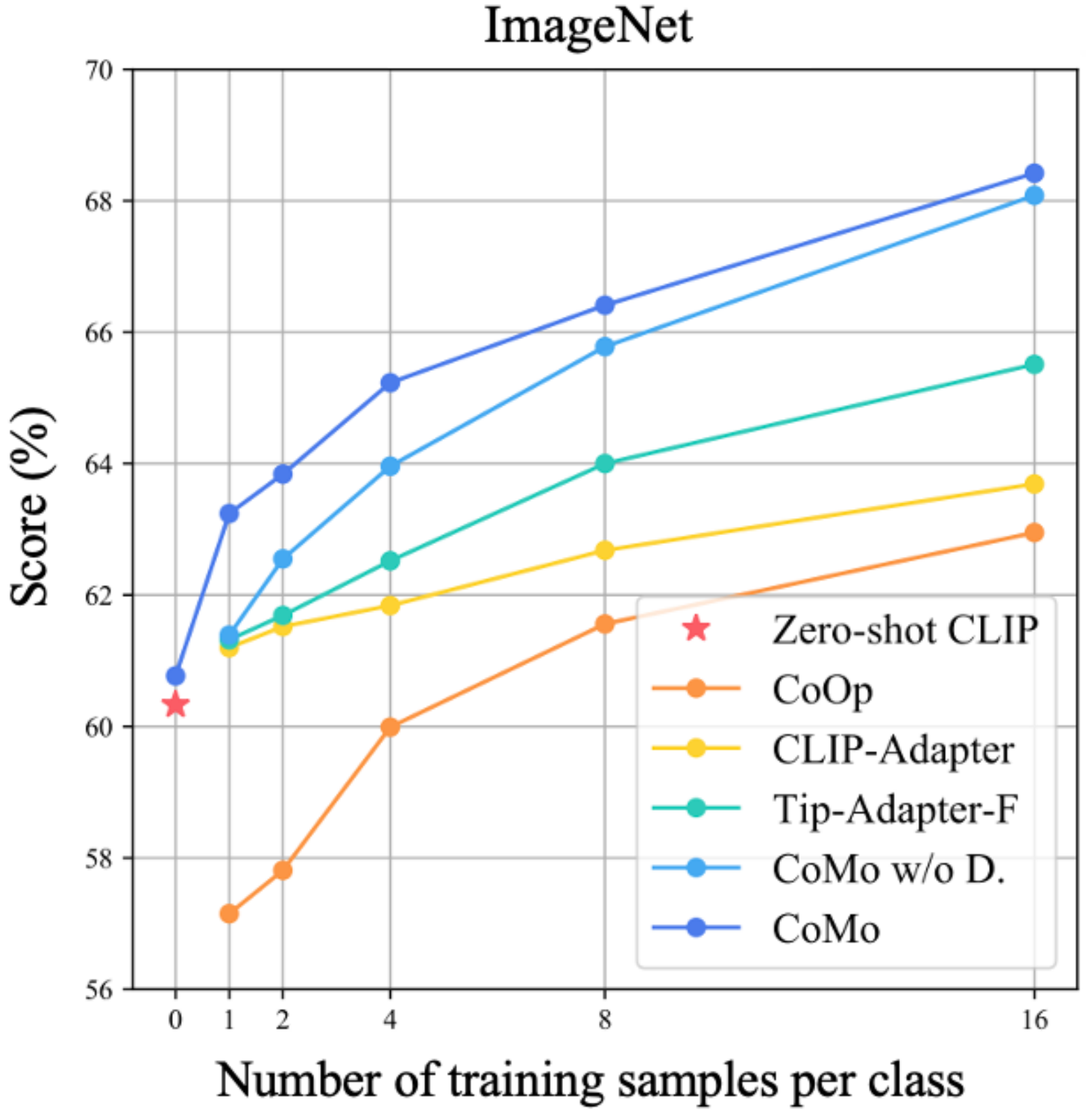}
\caption{\textbf{Performance (\%) Comparison on ImageNet.} We compare our CoMo with other methods for different few-shot settings, where ``CoMo w/o D.'' denotes CoMo without DALL-E's generated images.}
\label{fig:Imagenet Performance comparison}
\end{figure}
    
\section{Experiments}

\subsection{Settings}
\paragraph{Datasets.} We conduct few-shot experiments on 11 publicly available datasets: ImageNet~\cite{deng2009imagenet}, StandfordCars~\cite{krause20133d}, UCF101~\cite{soomro2012ucf101}, Caltech101~\cite{fei2004learning}, Flowers102~\cite{nilsback2008automated}, SUN397~\cite{xiao2010sun}, DTD~\cite{cimpoi2014describing}, EuroSAT~\cite{helber2019eurosat}, FGVCAircraft~\cite{maji2013fine}, OxfordPets~\cite{parkhi2012cats}, and Food101~\cite{bossard2014food}. We follow \cite{zhang2021tip} to train our CoMo with 1, 2, 4, 8, 16 shots and test on the full test set. As we could adopt DALL-E to generate training images in a zero-shot manner, we also report CoMo's zero-shot performance without the downstream few-shot training set.

\paragraph{Implementation.}
Our CoMo integrates the knowledge from pre-trained CLIP~\cite{radford2021learning}, DINO~\cite{Caron_2021_ICCV}, and DALL-E~\cite{pmlr-v139-ramesh21a}. For CLIP, we utilize ResNet-50~\cite{He_2016_CVPR} as the visual encoder and a transformer as the textual encoder. To align with the visual representation from CLIP, we also adopt DINO pre-trained upon ResNet-50. For DALL-E, we adopt different domain-specific textual templates as the input for different datasets, which correspond to the textual prompts for CLIP's textual encoder. During training, we only set the two kinds of keys in MK-Adapter to be learnable and utilize the data augmentation following Tip-Adapter. We train CoMo using batch size 64 only for 20 epochs, and adopt AdamW optimizer with the initial learning rate 0.0001 and a cosine scheduler.


\setlength{\tabcolsep}{2pt}
    \begin{table}[t]
    \centering
    \begin{tabular}{lcccccc}
    \toprule
        \multirow{1}*{Few-shot Setting} &0 &1 &2 &4 &8 &16\\
        \cmidrule(lr){1-1}
        \cmidrule(lr){2-7}
         CLIP &60.33 &-&-&-&-&-\\
         Linear-probe CLIP &- &22.17 & 31.90 &41.20 &49.52 &56.13\\
         CoOp &- &57.15 &57.81 &59.99 &61.56 &62.95\\
         CLIP-Adapter &- &61.20 &61.52 &61.84 &62.68 &63.59\\
         Tip-Adapter-F &- &61.32 &61.69 &62.52 &64.00 &65.51\\
        \midrule
         \textbf{CoMo w/o D.} &- & 61.39 & 62.55 &63.96 & 65.78 & 68.08 \\
         \textbf{CoMo} &\bf60.77 &\bf63.24 &\bf63.84 &\bf65.23 &\bf66.41 &\bf68.42 \\
         \bottomrule
    \end{tabular}
    \caption{\textbf{Quantative Performance (\%) Comparison on ImageNet.} We compare our CoMo with other methods for different few-shot settings, where ``CoMo w/o D.'' denotes CoMo without DALL-E's generated images.}
    \label{tab:imagenet performance comparison}
    \end{table}
    \setlength{\tabcolsep}{1.4pt}

\setlength{\tabcolsep}{3pt}
    \begin{table}[t]
    \centering
    \begin{tabular}{lcccc}
    \toprule
        \multirow{1}*{Models} &Epochs &Time &Accuracy &Gain\\
        \cmidrule(lr){1-1}
        \cmidrule(lr){2-5}
         Zero-shot CLIP & 0 & 0 & 60.33 & -\\
         Linear-probe CLIP & - & 13min & 56.13 &-4.20\\
         CoOp &200 &14h\ 40min &62.95 &+2.62\\
         CLIP-Adapter &200 &50min &63.59 &+3.26\\
         Tip-Adapter-F &20 &\textbf{5min} &65.51 &+5.18\\
         \textbf{CoMo} &20 &10min &\bf68.42 &\bf+8.09\\
         \bottomrule
    \end{tabular}
    \caption{\textbf{Efficiency Comparison on ImageNet.} We test the training time with a single A100 GPU under 16-shot setting.}
    \label{tab:time comparison}
    \end{table}
    \setlength{\tabcolsep}{1.4pt}
    
\begin{figure*}[t!]
    \centering
    \subfloat{\includegraphics[scale=0.225]{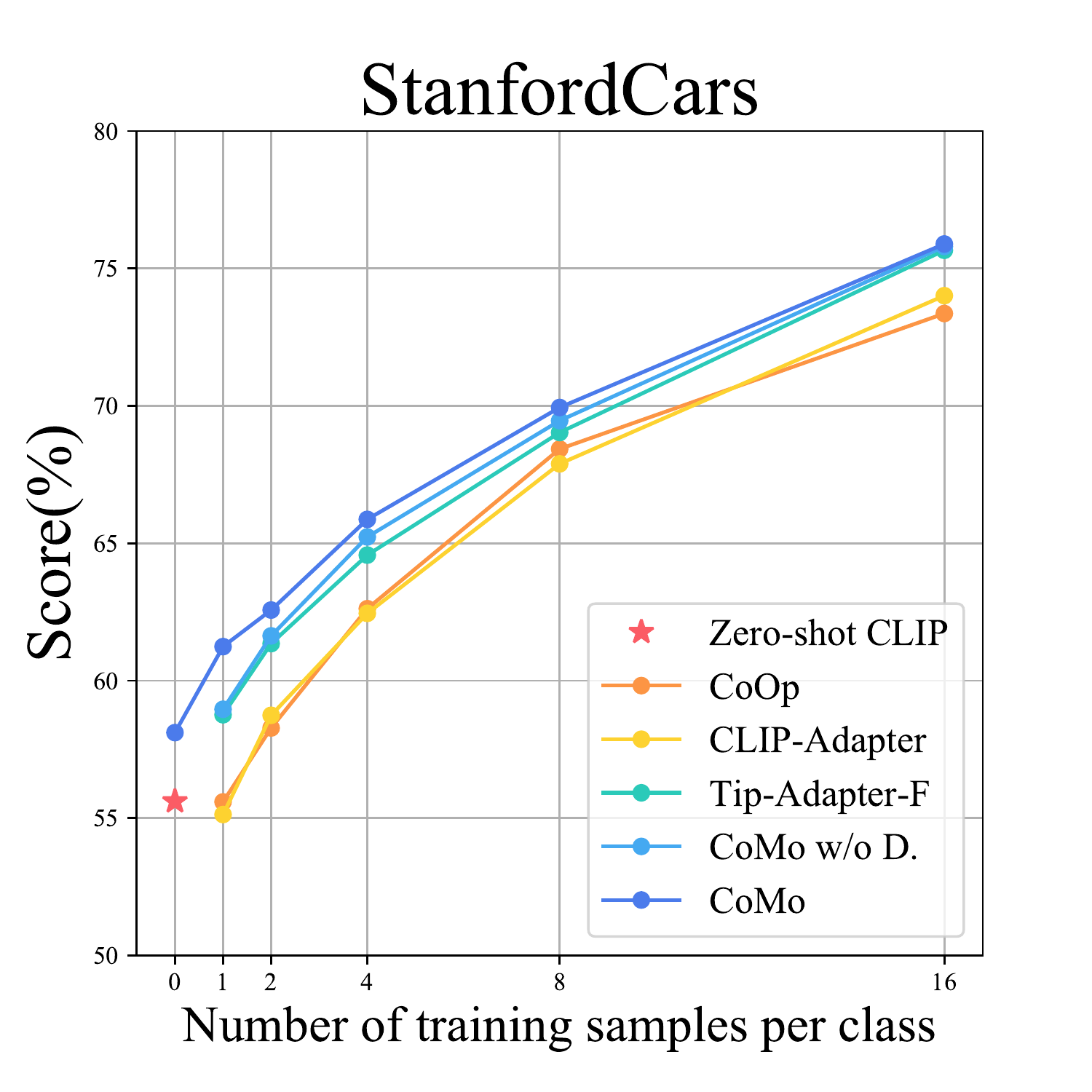}}
    \subfloat{\includegraphics[scale=0.225]{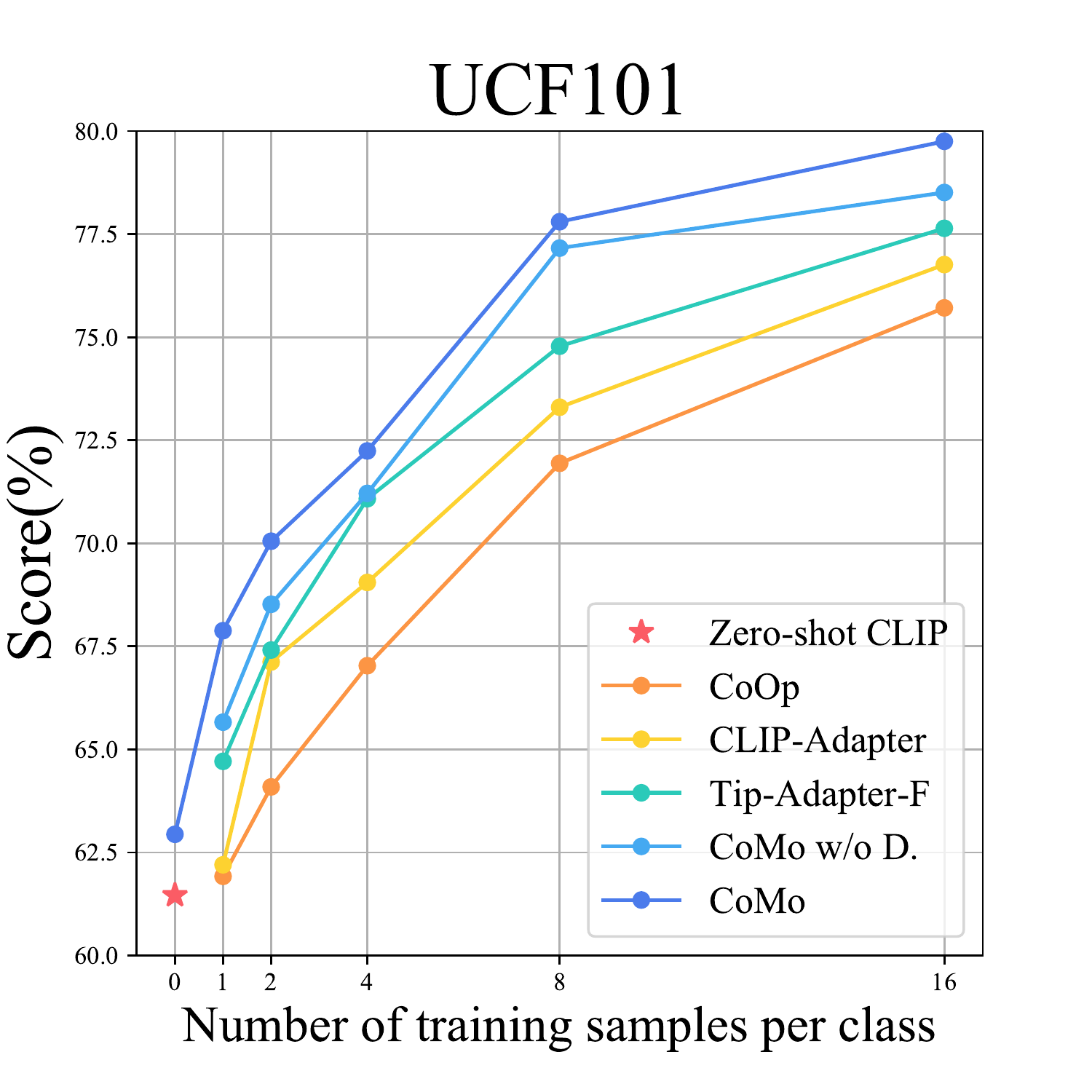}}
    \subfloat{\includegraphics[scale=0.225]{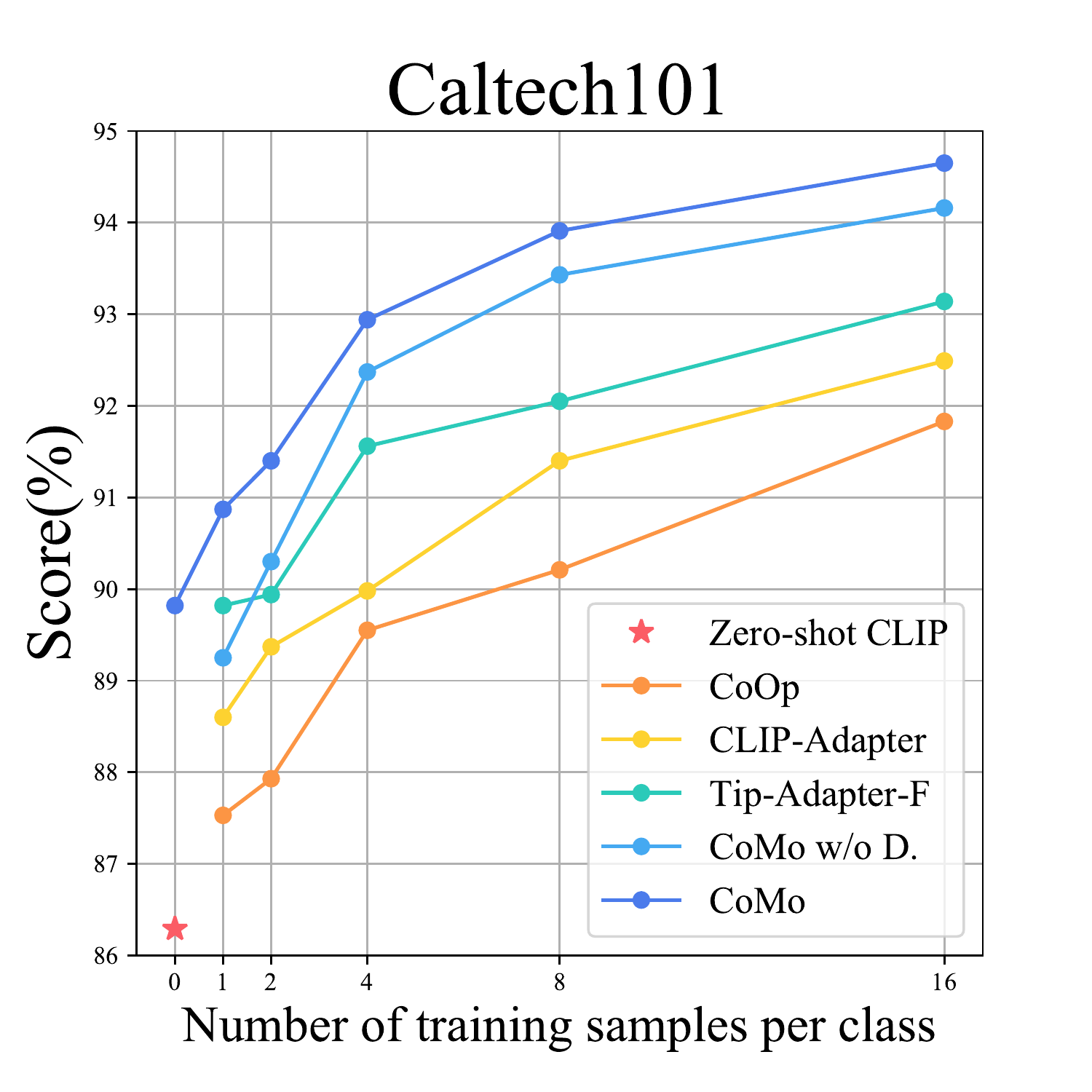}}
    \subfloat{\includegraphics[scale=0.225]{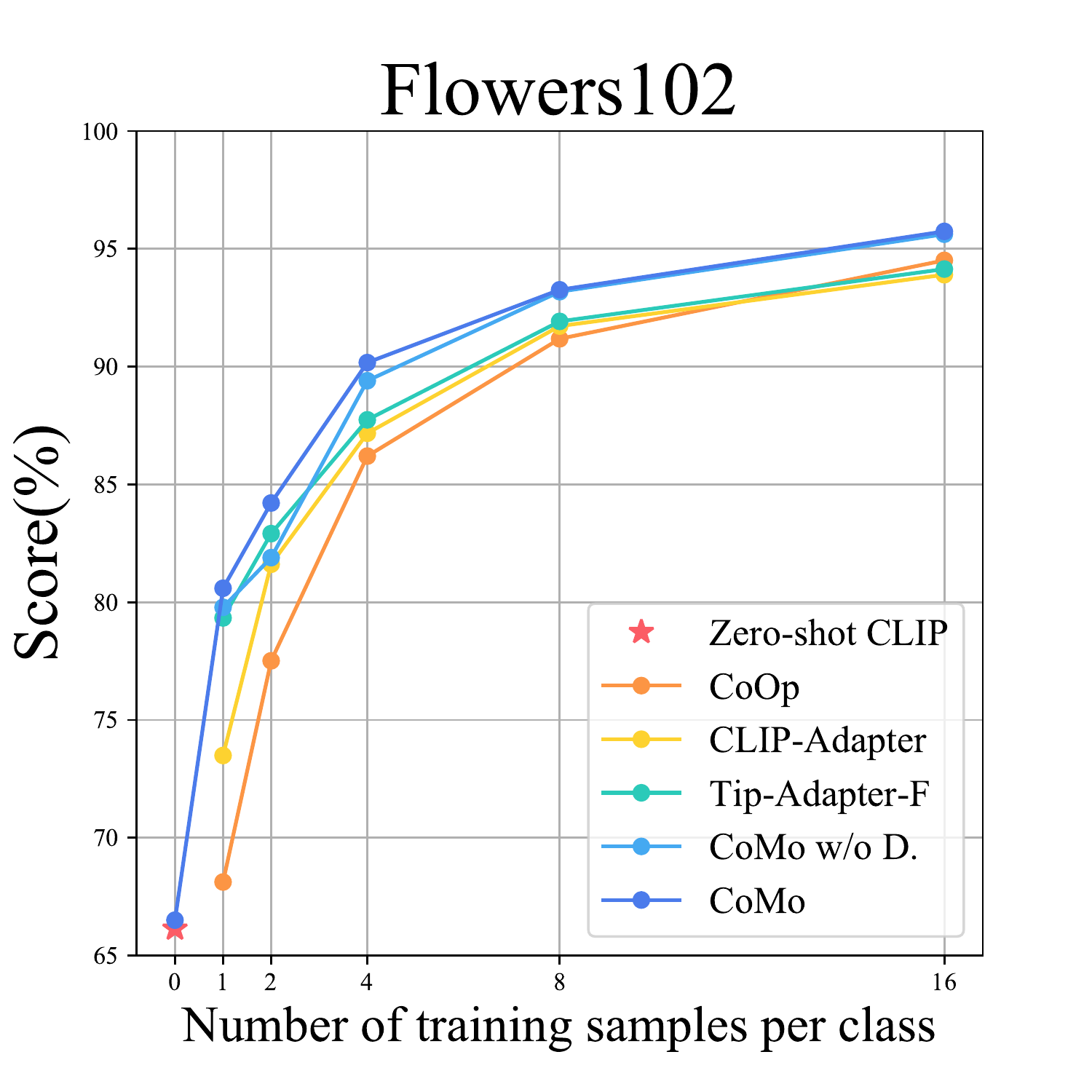}}
    \subfloat{\includegraphics[scale=0.225]{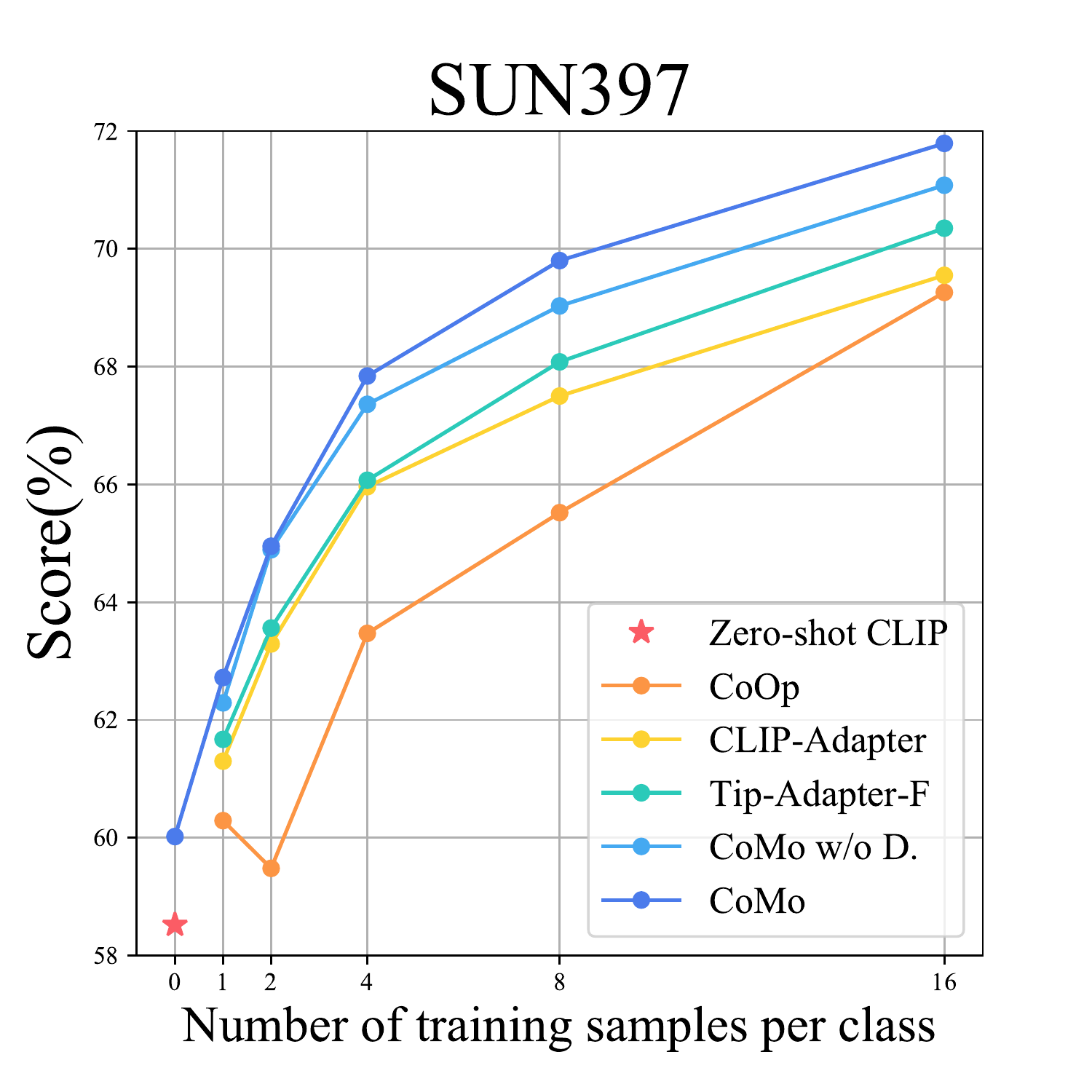}}
    \\
    \subfloat{\includegraphics[scale=0.225]{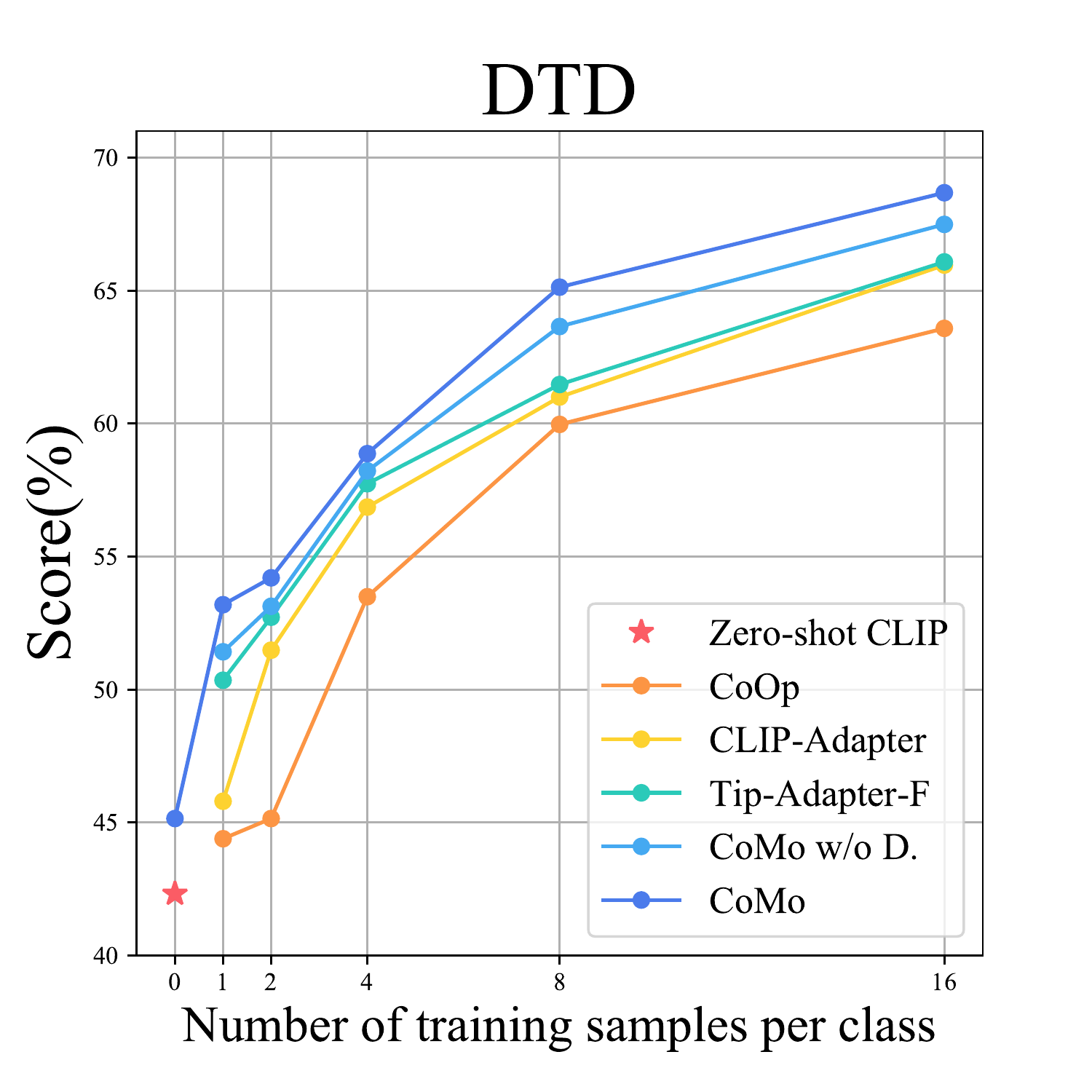}}
    \subfloat{\includegraphics[scale=0.225]{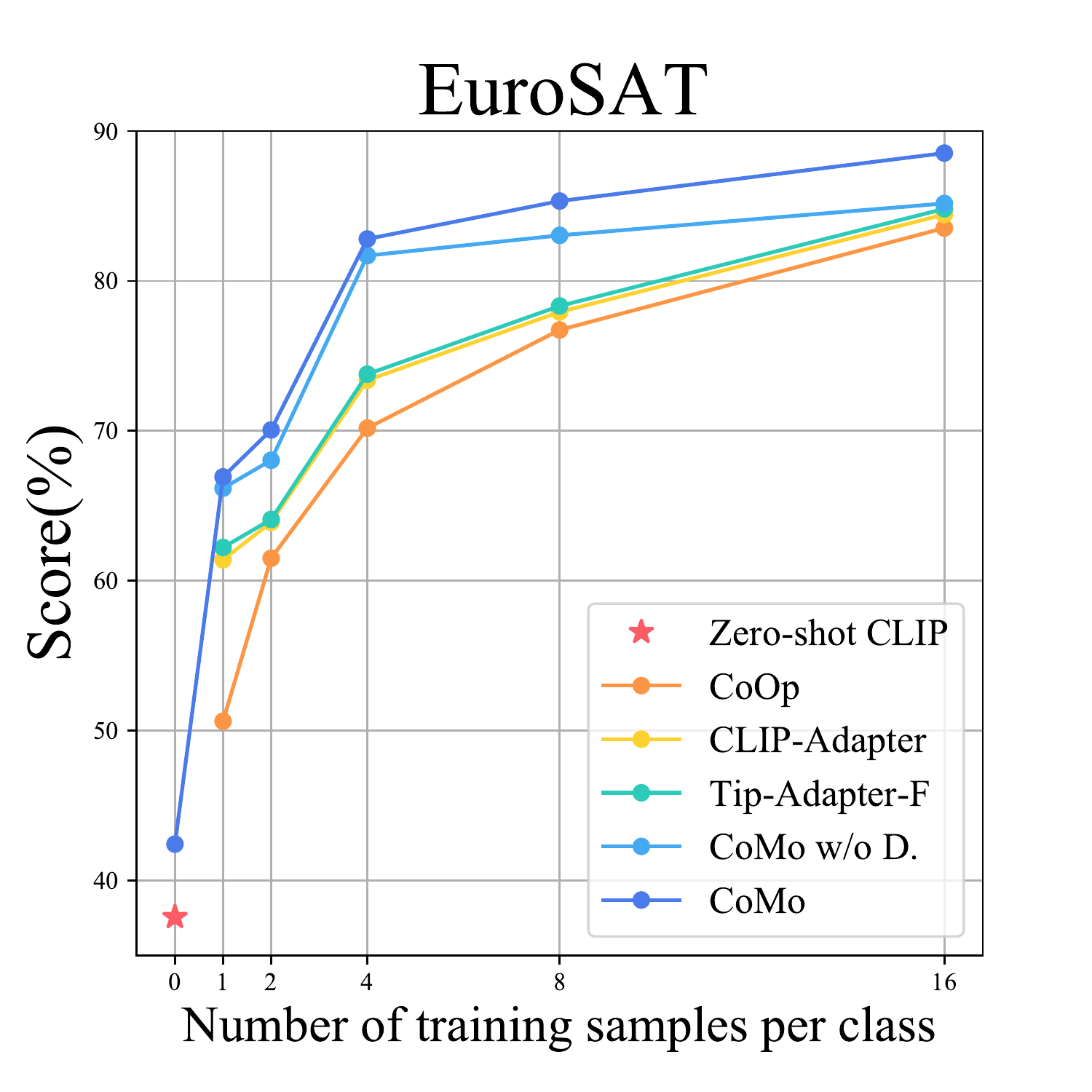}}
    \subfloat{\includegraphics[scale=0.225]{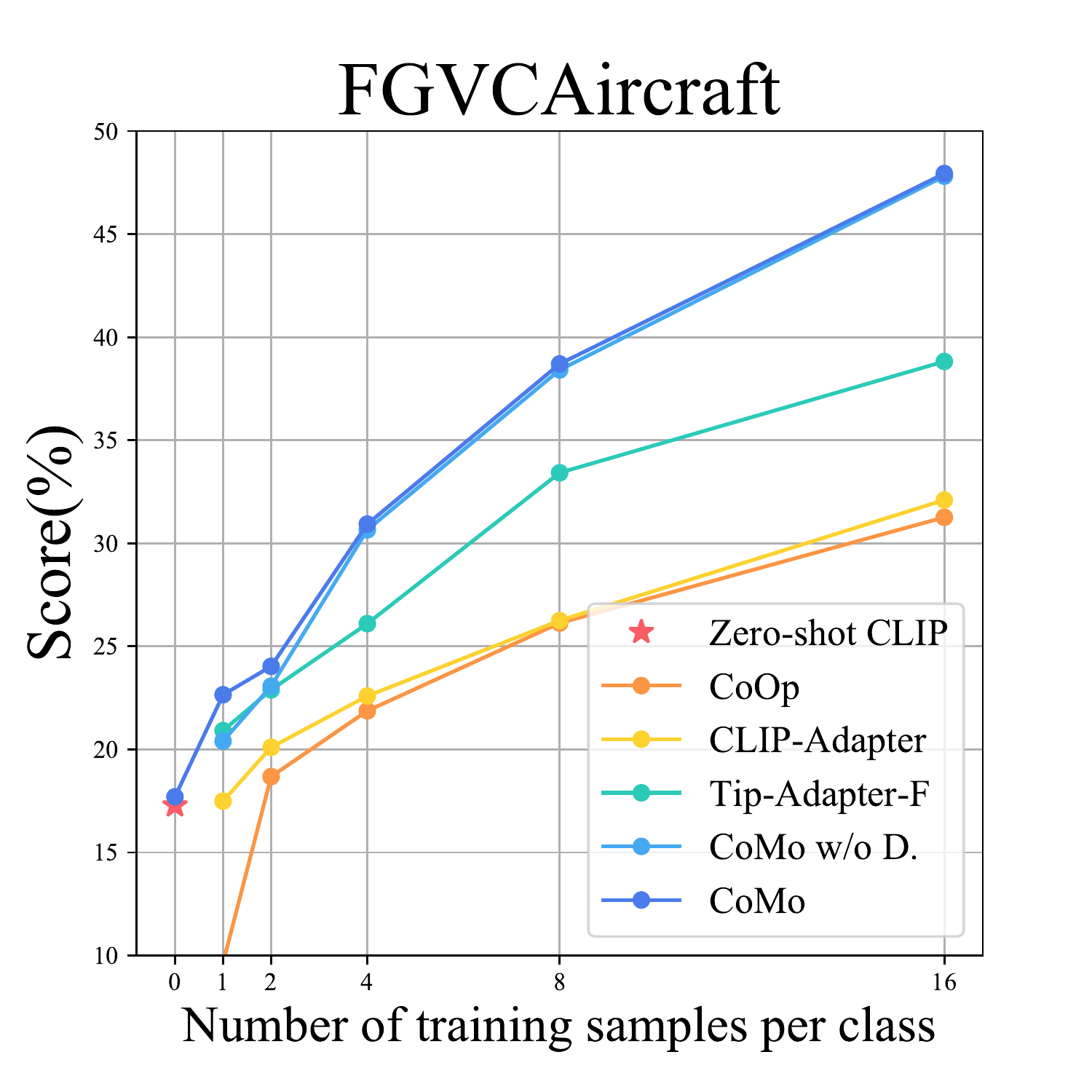}}
    \subfloat{\includegraphics[scale=0.225]{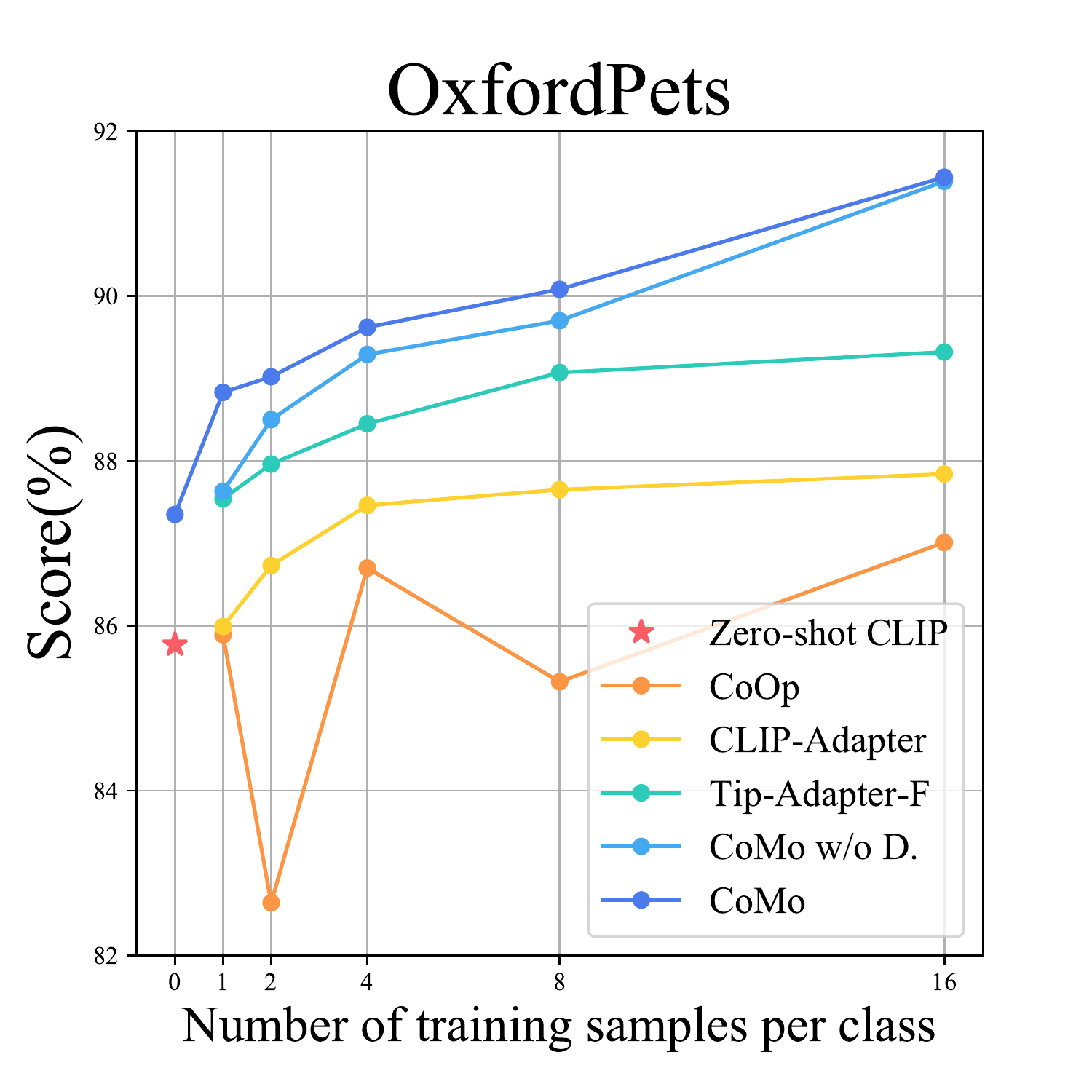}}
    \subfloat{\includegraphics[scale=0.225]{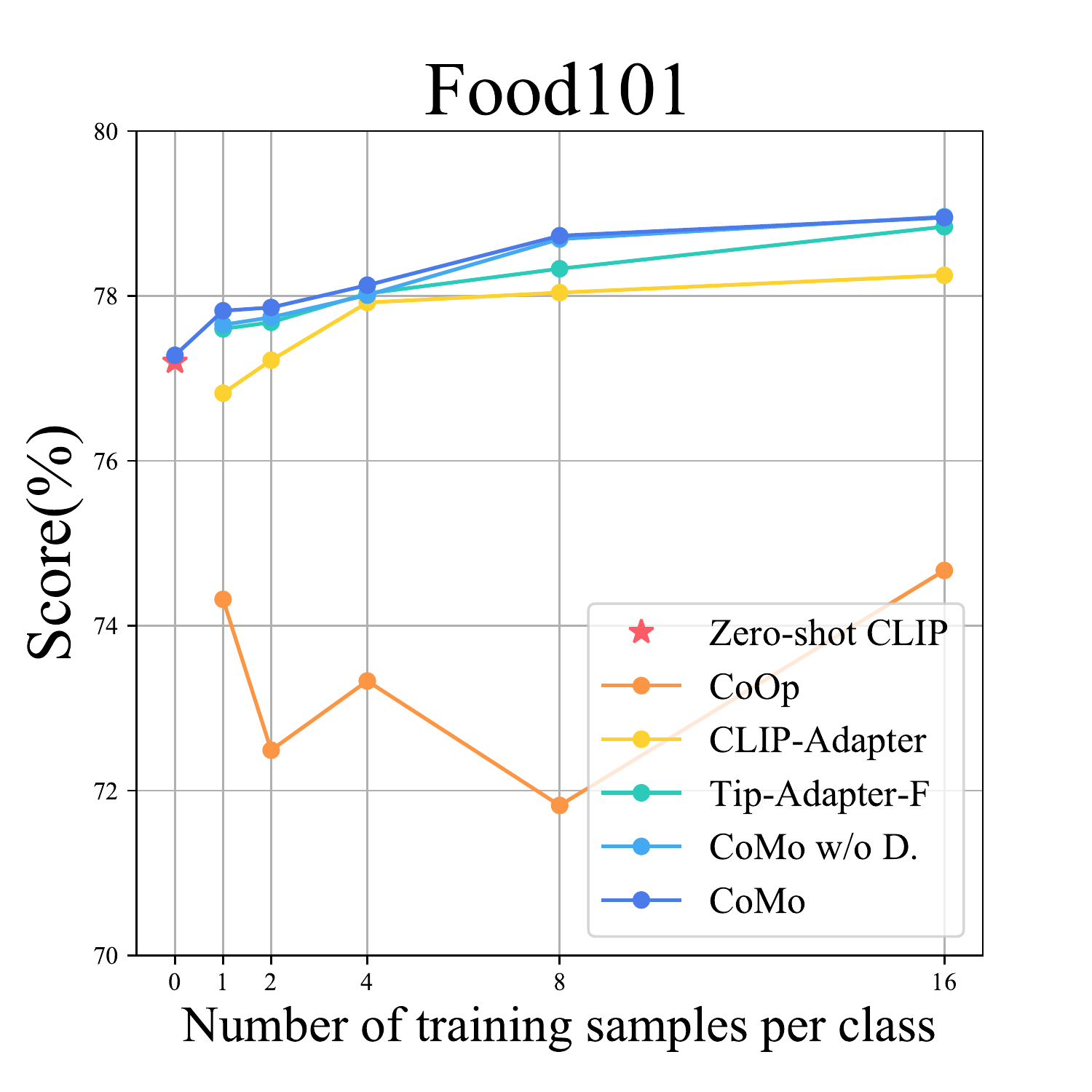}}
    \caption{\textbf{Performance (\%) Comparison on 10 Datasets.} Our method shows \textit{state-of-the-art} performance for all few-shot settings on different datasets.}
    \label{fig:other_dataset}
\end{figure*}
    
\subsection{Performance}

\paragraph{On ImageNet.}
We compare CoMo with other CLIP-based adaption methods on the most representative ImageNet: Linear-probe CLIP~\cite{radford2021learning}, CoOp~\cite{coop}, CLIP-Adapter~\cite{2110.04544}, and Tip-Adapter-F~\cite{zhang2021tip}. All these methods are based on the pre-trained CLIP~\cite{radford2021learning} with ResNet-50 visual encoders. 
As reported in Figure~\ref{fig:Imagenet Performance comparison} and Table~\ref{tab:imagenet performance comparison}, our CoMo surpasses all existing methods for different shot settings.
Remarkably, CoMo with 1 shot even outperforms the 8-shot Linear-probe CLIP and CoOp, and CoMo with 8 shots is better than all methods with 16 shots.
Since our method benefits from the zero-shot generation of DALL-E, we also compare the results without DALL-E, denoted as ``CoMo w/o D.'', which still remains \textit{state-of-the-art}. In Table~\ref{tab:time comparison}, we present the efficiency comparison of CoMo and other methods concerning training epochs and time. Our CoMo achieves the best performance-efficiency trade-off with 68.42\% accuracy and only 10 minutes training.




\paragraph{On Other Datasets.}
To further assess the robustness in different scenarios, we test CoMo on the extra 10 datasets in Figure \ref{fig:other_dataset}. For different semantic domains including real-world scenes, detailed textures, and satellite-captured landscapes, CoMo consistently shows leading performance and indicates excellent robustness via the collaboration of diverse knowledge. Also on some datasets, e.g., Caltech101 and OxfordPets, the zero-shot CoMo perform even comparably to other methods with 4 shots, demonstrating the effectiveness of zero-shot DALL-E for few-shot data expansion. 

\setlength{\tabcolsep}{8pt}
\begin{table}[t]
\begin{tabular}{lccc}
\toprule
\multirow{2}{*}{Datasets} & \textbf{Source} &\multicolumn{2}{c}{\textbf{Target}} \\
\cmidrule(lr){2-2} \cmidrule(lr){3-4} 
& ImageNet  & -V2 & -Sketch  \\ \midrule
Zero-shot CLIP  & 60.33  & 53.27 & 35.44\\
Linear-probe CLIP  & 56.13  & 45.61 & 19.13\\
CoOp & 62.95  & 54.58 & 31.04  \\
CLIP-Adapter &  {63.59}  &  {55.69} &  {35.68} \\
Tip-Adapter-F & 65.51  & 57.11 & 36.00 \\
\textbf{CoMo}            & \textbf{68.42} & \textbf{57.99} & \textbf{39.43}\\
\bottomrule
\end{tabular}
\caption{\textbf{Distribution Shift (\%) Comparison.} We train the models on ``Source'' dataset and test on ``target'' datasets.}
\label{tab:domain shifts}
\end{table}
\setlength{\tabcolsep}{1.4pt}

\paragraph{Distribution Shift.}
We further evaluate our CoMo's robustness to distribution shift by training on ``Source'' dataset and testing on ``Target'' datasets. In Table~\ref{tab:domain shifts}, we select the ``Source'' as ImageNe with the ``Target'' as ImageNet-V2~\cite{recht2019imagenet} and ImageNet-Sketch~\cite{hendrycks2021natural}. As shown, by incorporating diverse knowledge from pre-trained models, we achieve the best out-of-distribution performance on the two ``Target'' datasets.


\setlength{\tabcolsep}{3pt}
    \begin{table}[h]
    \begin{center}
    \begin{tabular}{cccccccc}
    \toprule\noalign{\smallskip}
    \multicolumn{3}{c}{{Pre-trained Models}} & \multicolumn{5}{c}{{Shot}}\\
    \noalign{\smallskip}
    \cmidrule(lr){1-3}
    \cmidrule(lr){4-8}
    \noalign{\smallskip}
    CLIP &DINO & DALL-E &1 &2 &4 &8 &16 \\
    \noalign{\smallskip}
    \cmidrule(lr){1-8}
    \noalign{\smallskip}
    \multicolumn{1}{c}{\checkmark} &{} &{} & 61.32 & 61.69 & 62.52 & 64.00 & 65.51 \\ 
    {} & \multicolumn{1}{c}{\checkmark} &{} & 34.14 & 34.44 & 40.47 & 44.32 & 53.27 \\
    \multicolumn{1}{c}{\checkmark} & \multicolumn{1}{c}{\checkmark} &{} & 61.39 & 62.55 &63.96 & 65.78 & 68.08 \\\cmidrule(lr){1-8}
    \multicolumn{1}{c}{\checkmark} &{} & \multicolumn{1}{c}{\checkmark} & 61.36 & 61.78 & 62.83 &64.04 &65.53\\ 
    {} & \multicolumn{1}{c}{\checkmark} & \multicolumn{1}{c}{\checkmark} & 34.13 & 34.44 & 41.12 & 45.01 & 53.63\\\cmidrule(lr){1-8}
    \multicolumn{1}{c}{\checkmark} & \multicolumn{1}{c}{\checkmark} & \multicolumn{1}{c}{\checkmark} &\bf63.24 &\bf63.84 &\bf65.23 &\bf66.41 &\bf68.42\\
    \bottomrule
    \end{tabular}
    \end{center}
    \caption{\textbf{Ablation Study (\%) of Collaboration Models.} We ensemble different pre-trained models on ImageNet.}
    \label{table:multi pretrain model ablation module}
    \end{table}
    \setlength{\tabcolsep}{1.4pt}

\subsection{Ablation study}

\paragraph{Collaboration Models.}
In Table \ref{table:multi pretrain model ablation module}, we explore how each pre-trained model contributes to the collaboration on different shots of ImageNet. Therein, ``CLIP'' denotes the zero-shot CLIP with MK-Adapter containing only CLIP's keys, and ``DINO'' denotes only the MK-Adapter with DINO's keys. As shown in the first three rows, the CLIP's language-contrastive knowledge performs stronger than DINO's vision-contrastive knowledge, which might benefit from millions of pre-training data. Their adaptive ensemble by MK-Adapter can bring larger improvement when the shot number increases. For the next two rows, DALL-E can boost both CLIP and DINO for nearly all shots with the generated synthetic images. The last row represents our final solution, CoMo that incorporates all three pre-trained models with the best performance for all shots.


\paragraph{Zero-shot Generation of DALL-E.}
We utilize DALL-E to generate synthetic images as the expanded few-shot training data. In Table~\ref{tab:dalle number}, we explore the best synthetic number $K'$ for each category of different shots on ImageNet. We observe that the larger $K'$ does not lead to better few-shot performance. As we adopt pre-trained CLIP to select the top-$K'$ generated images, which are scored by the similarities between CLIP-encoded images and category texts, the larger $K'$ would contain more low-quality images and adversely affect the MK-Adapter. Furthermore, the amount of expanded data is comparable to the original $K$ shots and thus preserves the characteristic of few-shot learning. 

\paragraph{Adaptive Ensemble.}
In Table~\ref{tab:dynamic method ablation study}, we ablate different ensemble methods of CLIP and DINO's predictions on ImageNet. The first two rows represent the MK-Adapter with one type of keys respectively for two pre-trained models without ensemble. Then, we adopt average and maximum pooling between the two predictions and ensemble the result with $p_{\text{ZS}}$. However, such naive integration without adaptive weights causes accuracy degradation. In the last three rows, we calculate the distribution similarities for adaptive ensemble and respectively select the three logits as the baseline. As shown, using $p_{\text{ZS}}$ as the distribution baseline performs the best, since $p_{\text{ZS}}$ itself shows strong transfer ability and can effectively suppress the wrong predictions of other logits.


\begin{figure}[t]
\centering
\includegraphics[width=1\linewidth]{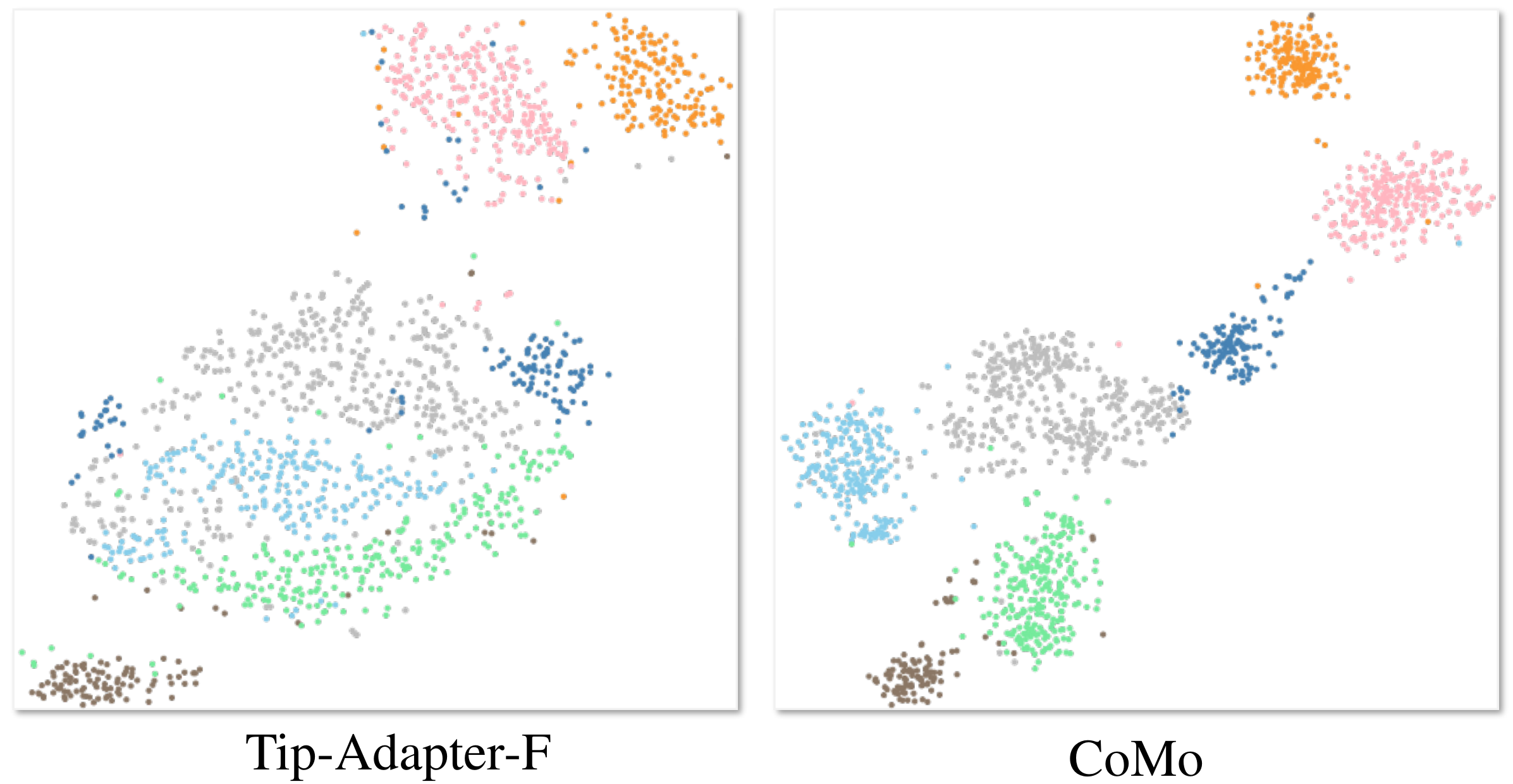}
\caption{\textbf{t-SNE Visualization.} Different colors represent different categories on 16-shot ImageNet.}
\label{fig:t-sne of como}
\end{figure}

\setlength{\tabcolsep}{7pt}
    \begin{table}[t]
    \centering
    \begin{tabular}{lccccc}
    \toprule
         \multicolumn{1}{c}{\multirow{2}*{Method}}&\multicolumn{5}{c}{{Shot}}\\
         \cmidrule(lr){2-6}
         &1 &2 &4 &8 &16\\
         \cmidrule(lr){1-6}
         CLIP & 61.36 & 61.78 & 62.83 &64.04 &65.53\\
         DINO & 34.13 & 34.44 & 41.12 & 45.01 & 53.63\\
         \cmidrule(lr){1-6}
         Average &60.70 &60.72 &60.99 &61.47 &61.97\\
         Maximum &61.64 &62.45 &62.95 &63.60 &64.97\\
         \cmidrule(lr){1-6}
         $p_{\text{CLIP}}$ Base. &62.36 &63.22 &64.11 &65.50 &67.40 \\
         $p_{\text{DINO}}$ Base. &62.61 &63.39 &64.31 &65.83 &67.73\\
         $p_{\text{ZS}}$ Base. & \bf63.24 &\bf63.84 &\bf65.23 &\bf66.41 &\bf68.42\\
         \bottomrule
    \end{tabular}
    \caption{\textbf{Ablation Study (\%) of Adaptive Ensemble.} We conduct different ensemble methods of Multi-Knowledge Adapter on ImageNet with different shots.}
    \label{tab:dynamic method ablation study}
    \end{table}
    \setlength{\tabcolsep}{1.4pt}

\paragraph{CLIP's Visual Encoders.}
We also conduct CoMo with different CLIP's visual encoders for comparison with other methods. As shown in Table~\ref{tab:Different vision backbone comparison}, CoMo consistently achieves leading performance with different visual backbones, indicating our generalizability to network architectures.

\begin{figure}[t]
\centering
\includegraphics[width=1\linewidth]{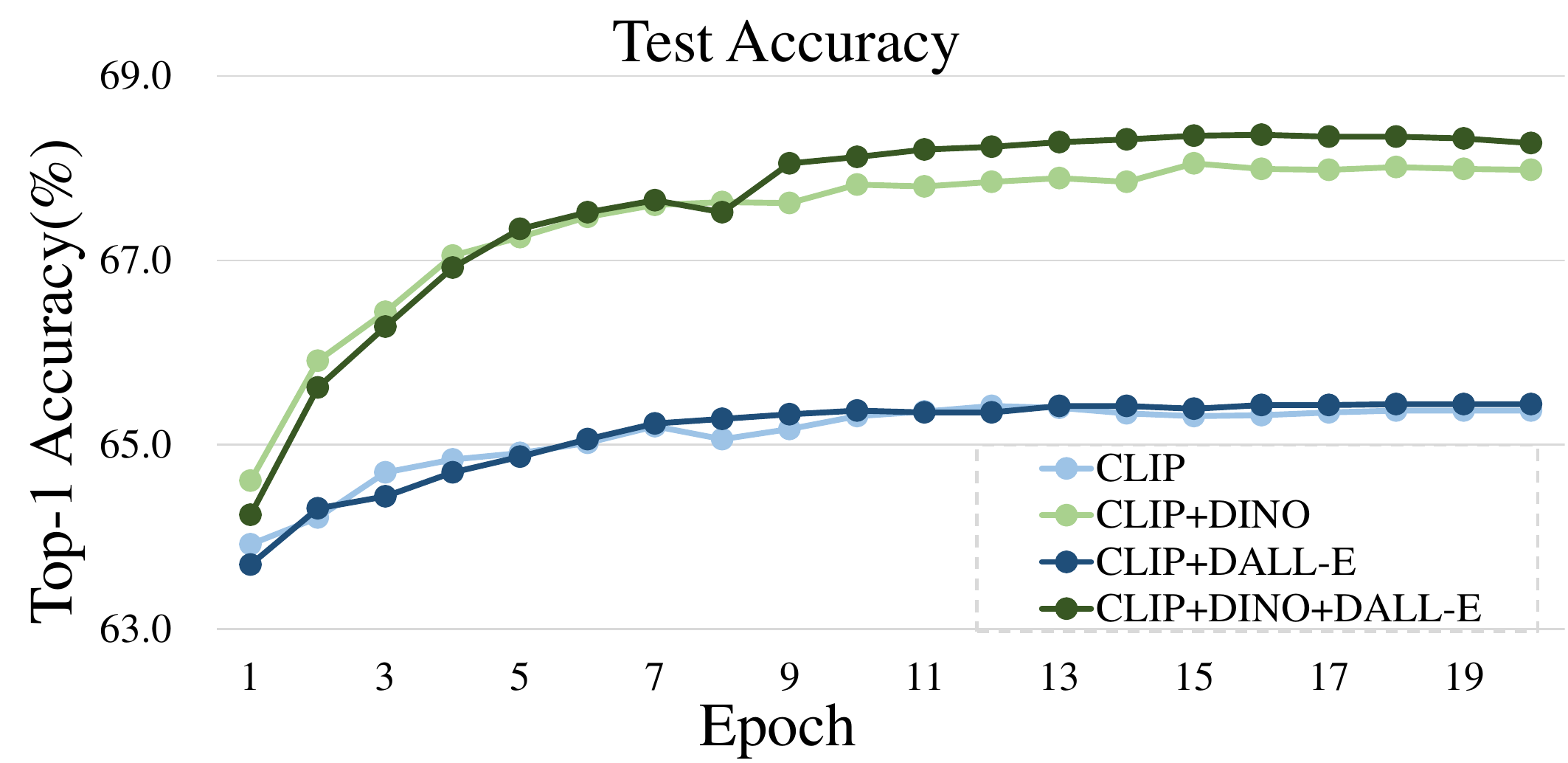}
\caption{\textbf{Learning Curves of Test Accuracy (\%)} for different combinations of pre-trained models during 20 epochs on 16-shot ImageNet.}
\label{fig:Imagenet training fig}
\end{figure}

\setlength{\tabcolsep}{8pt}
\begin{table}[t]
\begin{tabular}{lccccc}
\toprule
DALL-E & 1              & 2              & 4              & 8              & 16             \\ \midrule
1                          & 62.55          & 63.35          & 64.71          & 66.13          & 68.24          \\
2                          & 62.94          & \textbf{63.84} & 64.84          & 66.29          & 68.28          \\
4                          & \textbf{63.24} & 63.76          & 65.02          & 66.35          & \textbf{68.42} \\
8                          & 63.20          & 63.82          & \textbf{65.23} & 66.35          & 68.37          \\
16                         & 62.88          & 63.53          & 64.86          & \textbf{66.41} & 67.96          \\ \bottomrule
\end{tabular}
\caption{\textbf{Ablation Study (\%) of Zero-shot Generation of DALL-E.} We conduct CoMo with different number of generated images on ImageNet.}
\label{tab:dalle number}
\end{table}
\setlength{\tabcolsep}{1.4pt}

\setlength{\tabcolsep}{5pt}
    \begin{table}[t!]
    \centering
    \begin{tabular}{lcccc}
    \toprule
        \multirow{1}*{Models} &RN50 &RN101 &ViT-B/32 &ViT-B/16 \\
        \cmidrule(lr){1-1}
        \cmidrule(lr){2-5}
         Zero-shot CLIP &60.33 &62.53 &63.80 &68.73\\
         CoOp &62.95 &66.60 &66.85 &71.92 \\
         CLIP-Adapter &63.59 &65.39 &66.19 &71.13 \\
         Tip-Adapter-F&65.51 &68.56 &68.65 &73.69 \\
         \textbf{CoMo} &\bf68.42 &\bf70.58 &\bf70.68 &\bf74.48 \\
         \bottomrule
    \end{tabular}
    \caption{\textbf{Ablation Study (\%) of CLIP's Visual Encoders}. RN50 denotes ResNet-50, and ViT-B/32 denotes ViT-Base with patch size 32. We experiment on the 16-shot ImageNet.}
    \label{tab:Different vision backbone comparison}
    \vspace{-0.3cm}
    \end{table}
    \setlength{\tabcolsep}{1.4pt}

\subsection{Visualization}

\paragraph{t-SNE.} We present the t-SNE visualization of our CoMo and the second-best Tip-Adapter-F in Figure~\ref{fig:t-sne of como}. As shown, CoMo shows more contrastive distribution of category clusters than Tip-Adapter-F and can well mitigate some aliasing between simliar classes.

\paragraph{Learning Curves.} In Figure~\ref{fig:Imagenet training fig}, we visualize the 20-epoch learning curves of test accuracy on 16-shot ImageNet. Compared to the single CLIP, collaborating with DINO and DALL-E significantly improves the convergence speed and classification accuracy on test set.

\section{Conclusion}
We propose CoMo, a collaboration of models that comprehends diverse knowledge learned from different pre-training paradigms. Based on CLIP, we incorporate the language-generative DALL-E to expand the few-shot training data and adaptively fuse the vision-contrastive DINO via a Multi-Knowledge Adapter. By collaboration, CoMo achieves \textit{state-of-the-art} performance for few-shot learning on 11 datasets. Although CoMo has unified three types of pre-training, our future direction will focus on integrating more existing pre-trained knowledge, such as the mask-generated knowledge of MAE~\cite{He_2022_CVPR} and the 3D-contrastive knowledge of CrossPoint~\cite{afham2022crosspoint}.
{\small
\bibliography{aaai22}
}
\clearpage
\appendix

\section{Analysis by Other Metrics}

To evaluate our approach with other metrics, we employ the negative log-likelihood (NLL) and the area under the risk-coverage curve (AURC) to measure different ensemble models on 16-shot ImageNet~\cite{deng2009imagenet}. Lower NLL and AURC imply the better classification quality. In Table~\ref{tab:analyse model}, we observe that collaborating either DINO~\cite{Caron_2021_ICCV} or DALL-E~\cite{pmlr-v139-ramesh21a} with CLIP~\cite{radford2021learning} could improve the classification of each pre-trained model. However, different from CLIP, integrating DINO and DALL-E would slightly harm the performance. This might because the ImageNet-pre-trained DINO is unable to well recognize the synthetic images generated by DALL-E, but CLIP with much stronger transfer ability can be benefited from the expanded training set. In addition, the collaboration of all three pre-trained models contributes to the largest performance boost, which indicates the significance of their complementary characteristics.

\setlength{\tabcolsep}{5.5pt}
    \begin{table}[h]
    \centering
    \setcounter{table}{7}
    \begin{tabular}{cccccc}
    \toprule\noalign{\smallskip}
    \multicolumn{3}{c}{{Pre-trained Models}} & \multicolumn{3}{c}{{16-shot ImageNet}}\\
    \noalign{\smallskip}
    \cmidrule(lr){1-3}
    \cmidrule(lr){4-6}
    \noalign{\smallskip}
    CLIP &DINO & DALL-E &NLL $\downarrow$ &AURC $\downarrow$ &Acc. $\uparrow$ \\
    \noalign{\smallskip}
    \cmidrule(lr){1-6}
    \noalign{\smallskip}
    \multicolumn{1}{c}{\checkmark} & {} & {} &1.29 &141.9 &65.51\\ 
    {} & \multicolumn{1}{c}{\checkmark} & {} &2.96 &245.0 &53.27\\
    \multicolumn{1}{c}{\checkmark} & \multicolumn{1}{c}{\checkmark} & {} &1.27 &132.0 & 68.08\\
    \cmidrule(lr){1-6}
    \multicolumn{1}{c}{\checkmark} & {} & \multicolumn{1}{c}{\checkmark}  &1.29 & 141.6 & 65.53\\ 
    {} & \multicolumn{1}{c}{\checkmark} & \multicolumn{1}{c}{\checkmark} &3.45 &248.6 &53.63\\
    \cmidrule(lr){1-6}
    \multicolumn{1}{c}{\checkmark} & \multicolumn{1}{c}{\checkmark} & \multicolumn{1}{c}{\checkmark} &\bf1.22 &\bf121.2 &\bf68.42\\
    \bottomrule
    \end{tabular}
    \caption{\textbf{NLL and AURC ($\times10^3$) of Different Models.} NLL, AURC, and Acc. (\%) denote the negative log-likelihood, area under the risk-coverage curve, and classification accuracy, respectively.}
    \label{tab:analyse model}
    \end{table}
    \setlength{\tabcolsep}{1.4pt}

\section{Additional Ablation Study}

\paragraph{Zero-shot CoMo.}
As we leverage the pre-trained DALL-E to generate the supplementary few-shot training set in a zero-shot manner, our CoMo can be evaluated under zero-shot settings the same as CLIP, for which none of the human-annotated training images is given. In Table~\ref{tab:Zero shot ablation study}, we report the best generated image number $K'$ of DALL-E for zero-shot CoMo. The number ``0'' denotes Zero-shot CLIP. For different datasets, the best number varies ranging from 1$\sim$16, and the larger number normally cannot get the better result, probably due to the low-quality synthetic images. On Caltech101~\cite{fei2004learning} and EuroSAT~\cite{helber2019eurosat}, zero-shot CoMo largely surpasses CLIP by +3.53\% and +4.87\%, indicating our superiority under zero-shot settings.

\paragraph{Hyperparameter $\beta$.}
In Formula 5 and 6, we utilize a non-linear modulator $\varphi(x) = \exp(-\beta\cdot(1-x))$ for the affinity matrix of CLIP and DINO in the MK-Adapter, where $\beta$ controls the matrix sharpness. In Table~\ref{tab:hyperparameter ablation}, we experiment CoMo with different $\beta$ on 16-shot ImageNet and observe 0.6 performs the best.

\section{Additional Visualization}

In Figure~\ref{fig:image_comparison}, we visualize the synthetic images generated by DALL-E on ImageNet~\cite{deng2009imagenet}, OxfordPets~\cite{parkhi2012cats}, and Caltech101~\cite{fei2004learning} datasets. As shown, benefited from the pre-trained DALL-E, the generated images can well highlight the semantics of target category and effectively expand the few-shot training set in low-data regimes.

\setlength{\tabcolsep}{1pt}
    \begin{table}[h]
    \begin{center}
    \begin{tabular}{c|cccccc}
    \toprule
    \text{Sharpness $\beta$\ } & 0.4 & 0.5 & 0.6 & 0.7 & 0.8 & 1.0 \\
    \cmidrule(lr){1-1} \cmidrule(lr){2-7}
     CoMo &\ \ 68.29\ \ &\ \ 68.38\ \ &\ \ \bf68.42\ \ &\ \ 68.36\ \ &\ \ 68.33\ \ &\ \ 68.29\ \ \\
    \bottomrule
    \end{tabular}
    \end{center}
    \caption{\textbf{Ablation Study (\%) of Hyperparameter $\beta$.}}
    \label{tab:hyperparameter ablation}
    \end{table}
    \setlength{\tabcolsep}{1pt}

\setlength{\tabcolsep}{9pt}
    \begin{table*}[t]
    \centering
    \begin{tabular}{cccccccccccc}
    \toprule
        {DALL-E} & \rotatebox{90}{ImageNet} & \rotatebox{90}{Caltech101} & \rotatebox{90}{Flower102} & \rotatebox{90}{Food101} & \rotatebox{90}{DTD} & \rotatebox{90}{EuroSAT} & \rotatebox{90}{OxfordPets} & \rotatebox{90}{SUN397} & \rotatebox{90}{StanfordCars} & \rotatebox{90}{UCF101} & \rotatebox{90}{FGVCAircraft}\\
        \cmidrule(lr){1-1}
        \cmidrule(lr){2-12}
        0 & 60.33 & 86.29 & 66.14 & 77.20 & 42.32 & 37.56 & 85.77 & 58.52 & 55.61 & 61.46 & 17.28\\
        1 & 60.61 & 89.05 & 66.34 & 77.24 & 44.98 & \bf42.43 & 87.27 & \bf60.62 & 57.18 & 62.20 & \bf17.70\\
        2 & 60.76 & 88.72 & \bf66.50 & 77.27 & 45.04 & 40.58 & 86.89 & 59.61 & 56.86 & 62.54 & 17.58\\
        4 & \bf60.77 & 89.13 & 66.26 & 77.24 & \bf45.15 & 40.01 & 87.16 & 59.94 & 56.97 & 62.44 & \bf17.70\\
        8 & 60.75 & 89.37 & 66.46 & \bf77.28 & 44.86 & 40.68 & \bf87.35 & 59.97 & 57.12 & 62.41 & 17.64\\
        16 & 60.74 & \bf89.82 & 66.38 & 77.27 & 44.86 & 39.81 & 87.00 & 59.73 & \bf58.11 & \bf62.94 & 17.49\\
        \bottomrule
    \end{tabular}
    \caption{\textbf{Ablation study (\%) of Zero-shot CoMo via DALL-E on Different Dataset.} We leverage DALL-E to generate different numbers of synthetic images for zero-shot recognition.}
    \label{tab:Zero shot ablation study}
    \end{table*}
    \setlength{\tabcolsep}{1.4pt}
\begin{figure*}[t]
\centering
\setcounter{figure}{7}
\includegraphics[width=1\linewidth]{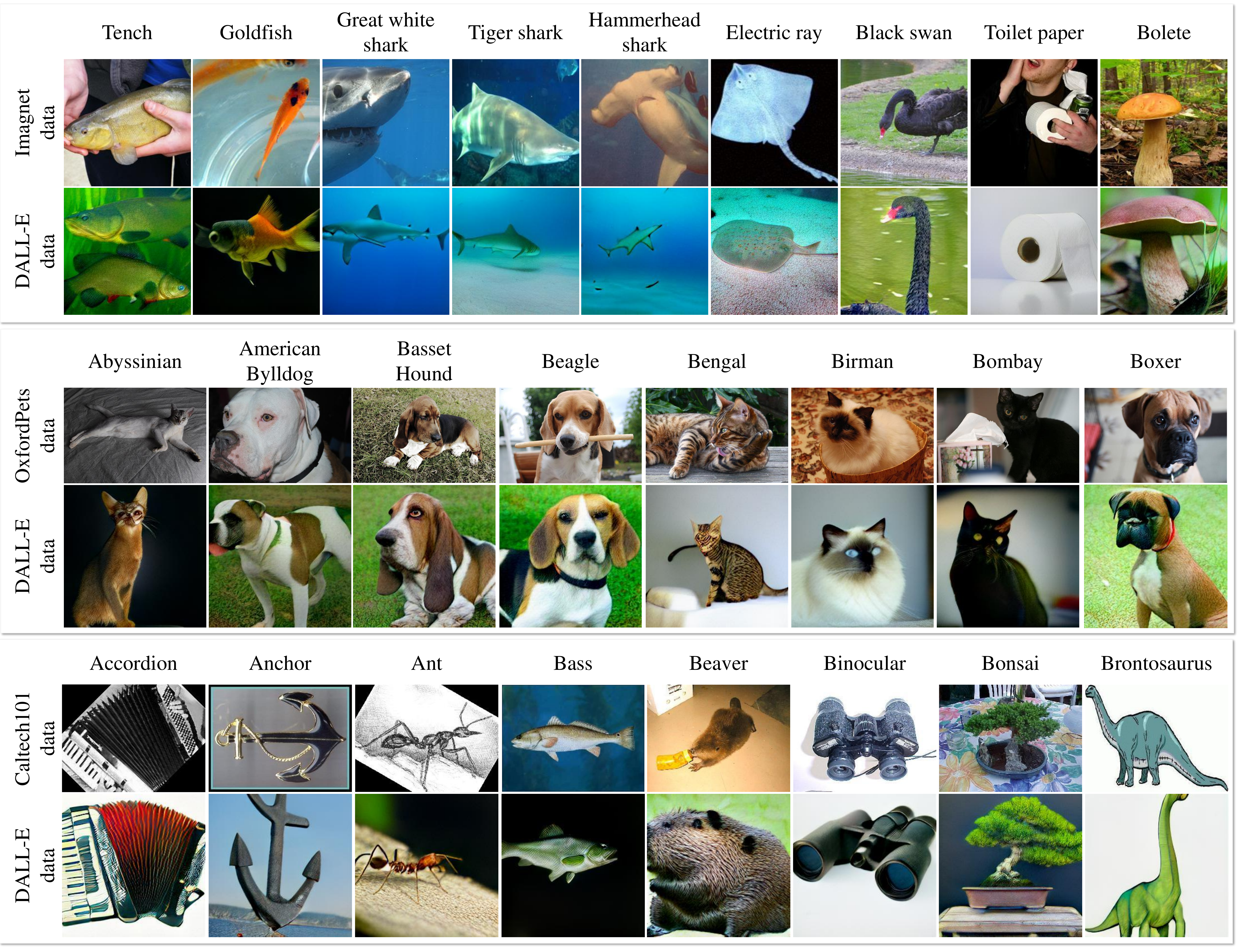}
\caption{\textbf{Visualizations of Generated Images by DALL-E.}}
\label{fig:image_comparison}
\end{figure*}

\end{document}